# Multirobot Cliff Climbing on Low-Gravity Environments

Himangshu Kalita
Space and Terrestrial Robotics Exploration Laboratory
School of Energy, Matter and Transport Engineering
Arizona State University
Tempe, Arizona, United States
hkalita@asu.edu

Jekan Thangavelautham
Space and Terrestrial Robotics Exploration Laboratory
School of Earth and Space Exploration
Arizona State University
Tempe, Arizona, United States
jekan@asu.edu

*Abstract*— Exploration of extreme environments, including caves, canyons and cliffs on low-gravity surfaces such as the Moon, Mars and asteroids can provide insight into the geological history of the solar system, origins of water, life and prospect for future habitation and resource exploitation. Current methods of exploration utilize large rovers that are unsuitable for exploring these extreme environments. In this work, we analyze the feasibility of small, low-cost, reconfigurable multirobot systems to climb steep cliffs and canyon walls. Each robot is a 30-cm sphere covered in microspines for gripping onto rugged surfaces and attaches to several robots using a spring-tether. Even if one robot were to slip and fall, the system would be held up with multiple attachment points much like a professional alpine climber. We analyzed and performed detailed simulations of the design configuration space to identify an optimal system design that trades-off climbing performance with risk of falling. Our results show that with increased number of robots, climbs can be performed faster (through parallelism) and with less risk of falling. The results show a pathway towards demonstration of the system on real robots.

*Keywords—climbing, extreme environment, multirobot system, adaptation, reconfigurability.*

## I. INTRODUCTION

The surfaces of the Moon and Mars have been explored with wheeled robots that house state-of-the-art science laboratories. These robots have provided insight into the geology and geo-history of these bodies [1][2]. However, these wheeled robots are large, in the order of 200 to 800 kg or more and are unable to explore extreme environments such as caves, canyons, cliffs and craters walls. This is due to inherent limitations of their large size, and challenges in motion planning and control on inclined natural surfaces. Rapid advancement in lightweight structural materials, miniaturization of electronics, sensors and actuators has enabled small, low-mass and low-cost platforms that can hop, perform short flights and roll. Use of Guidance, Navigation and Control (GNC) devices such as reaction-wheels, inertial measurements units and propulsion enable unprecedented mobility in precarious, low-gravity surface conditions.

In this paper, we propose use of spherical robots called SphereX for extreme environment climbing. These robots would be covered in a suitable gripping skin embedded with arrays of microspines that enables these robots to grasp onto rough terrain and rest on precarious/sloped surfaces. A single robot trying to hop and grip onto a sloped terrain may slip and fall. However, a multirobot system can work cooperatively by being interlinked using spring-tethers and work much like a team of mountaineers to systematically climb a slope. Each robot is secured to a slope using spiny gripping actuators, and one by one each robot moves upwards by crawling, rolling or hopping up the slope. If any one of the robots loses grip, slips or falls, the remaining robots will be holding it up as they are anchored [5]. We present dynamics and control simulations for such an autonomous multirobot system that cooperates to climb sloped natural terrains.

Multirobot systems hold great potential for exploring cliffs and rugged surfaces on planetary environments. Recent evidence suggests that water flowed down the faces of several Martian cliffs as seen in high-resolution images acquired by the Mars Reconnaissance Orbiter Camera [2]. Rappelling down and getting up close to these slopes is not possible using conventional wheeled, legged or rolling robots but can be achieved using the proposed multirobot climbing system. Moreover, on milligravity surfaces such as asteroids, hopping and flying is simple and uses negligible propellant. However, the gravity varies throughout the surface and too much thrust can result in the robot hopping off the asteroid. The proposed multirobot system with robots anchored to the surface is a viable mobility system in such milligravity surfaces as the anchored robots keeps the entire system secure.

## II. RELATED WORK AND MOTIVATION

A multitude of robotic systems have been proposed for extreme-environment mobility and exploration but climbing sloped natural terrains has always been a major challenge. To achieve exploration of sloped terrains much work has been done on developing tethered, legged and wheeled robotic systems. For exploring volcanoes, Dante II was developed which is an eight-legged walking rover with a tethered rappelling mobility system [6]. Teamed Robots for Exploration and Science on Steep Areas (TRESSA) is a dual-tethered robotic system used for climbing steep cliff faces with slopes varying from 50 to 90 degrees [7]. NASA JPL successfully demonstrated accessing 90 degree vertical cliffs and collecting samples using the Axel platform which is a two-wheeled rover tethered to its host platform [8]. Another example is the All-Terrain Hex-Limbed Extra-Terrestrial Explorer (ATHELETE) rover developed by NASA JPL which has 6-DOF limbs, each attached with a 1-DOF wheel [9].

Several other robotic systems have been developed that uses friction, suction cups, magnets and sticky adhesives to climb sloped terrains. The Legged Excursion Mechanical Utility Rover (LEMUR IIb) developed by NASA JPL and Stanford is a four-limbed robot capable of free-climbing vertical rocky surfaces, urban rubble piles, sandy terrains and roads using only friction at contact points [10]. Stickybot developed at Stanford employs several design principles adapted from the gecko lizard like hierarchial compliance, directional adhesion and force control to climb smooth surfaces at very low speeds [11]. Spinybot II can climb a wide variety of hard, outdoor surfaces including concrete, stucco, brick and sandstone by employing arrays of microspines that catch on surface irregularities [12]. The Robots in Scansorial Environments (RiSE) is a new class of vertical climbing robots that can climb a variety of human-made and natural surfaces employing a combination of biologically inspired attachments, dynamic adhesion and microspines [13]. NASA JPL has also developed an anchoring foot mechanism for sampling on the surface of near Earth asteroids using microspines that can withstand forces greater than 100 N on natural rock and has proposed to use it on the Asteroid Retrieval Mission (ARM) [14]. The examples described are all single robots. Control of multiple robots have posed challenges, however our previous work utilizing artificial neural networks show the potential benefits to multirobot cooperation in solving difficult tasks [22],[23],[24].

For the proposed multirobot system, the motivation is taken from proven methods used by alpinists to climb mountains. These mountaineers use ice axes and crampons to grip on the surface and climb steep mountain slopes as shown in Fig. 1. The use of legs and hands provide four contact points to the sloped surface. The systematic climbing approach is redundant in nature as even when each attempt to grip onto a higher location fails, the climber is still secure with his feet and one hand gripping tightly onto the slope. Inspired by mountaineers, our proposed approach is a systems solution to address the challenge of off-world climbing. The system utilizes multiple SphereX robots that are interlinked with spring-tether. Each robot is equipped with an array of microspines to grip on the rough surface.

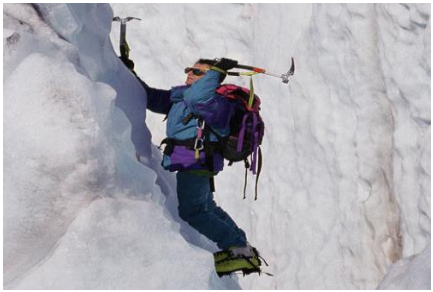

Fig. 1: Mountain climbers use multiple contact points to climb and stay anchored to a slope.

## III. SYSTEM DESIGN

For climbing inclined natural slopes, the multirobot system should have multiple contact points, subsystem redundancy and have a well-balanced mass. The fundamental issues involved in developing such a system includes hardware design, gripping, sensing, planning and control [15].

### A. Hardware Design

The multirobot system is designed to move in vertical/inclined natural slopes. The system consists of four identical spherical robots interlinked together with spring tethers in an "X" configuration. Each robot has a mass of 3 kg and diameter 0.3 m. Fig. 2 shows the internal and external views of each SphereX robot. The lower half of the sphere contains the power and propulsion system, with storage tanks for fuel and oxidizer connected to the main thruster. It also contains a 3-axis reaction wheel system for maintaining roll, pitch and yaw angles and angular velocities along x, y, and z axis. The propulsion unit provides thrust along the +z axis and the reaction wheel system controls the attitude and angular velocity of the robot that enables it to perform ballistic hop. Next is the Lithium Thionyl Chloride batteries with specific energy of 500 Wh/kg arranged in a circle. An alternative to batteries are PEM fuel cells. PEM fuel cells are especially compelling as techniques have been developed to achieve high specific energy using solid-state fuel storage systems that promise 2,000 Wh/kg [16][17][18]. However, PEM fuel cells require development for a field system in contrast to lithium thionyl chloride that has already been demonstrated on Mars.

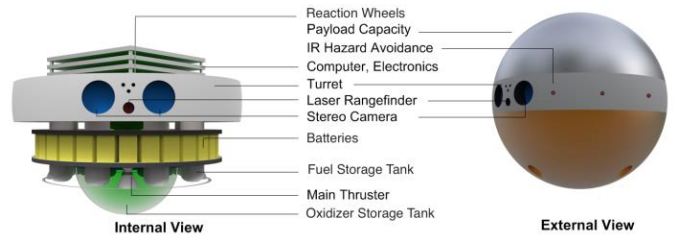

Fig. 2: Internal and External view of the SphereX Robot

For sensing, planning and control, a pair of stereo cameras and a laser range finder is mounted to each robot and they roll on a turret. This enables the robot to take panoramic pictures and scan the environment without having to move using the propulsion system. Above the turret are two computer boards, IMU and IO-expansion boards, in addition to a power board. The volume above the electronics is reserved for climbing mechanism and payload of up to 1 kg [3].

Apart from the proposed propulsion subsystem, all the other hardware components can be readily assembled using Commercial off-the-shelf (COTS) components. The proposed propulsion system uses RP-1 as the fuel and $H_2O_2$ as the oxidizer. Table I shows the mass budget for a single robot.

TABLE I. SPHEREX ROBOT MASS BUDGET

| Subsytem | Mass (kg) |
|---|---|
| Computer, Comms, Electronics | 0.2 |
| Power | 0.3 |
| Stereo Camera, Laser Rangefinder | 0.3 |
| Propulsion | 1 |
| ADCS | 0.4 |
| Climbing Payload | 0.8 |
| Total | 3 |

*B. Control of Each Robot*

System mobility is achieved by enabling each robot to perform ballistic hops. The sequence at which each robot performs ballistic hops will be covered in the planning section. This section descibes the controls approach to perform ballistic hopping for each individual robot using the propulsion unit and the 3-axis reaction wheel system [4].

The forces acting on each robot during ballistic hop are the thrust generated by the propulsion unit along +z axis with gravity pointing along the –z axis. The 3-axis reaction wheel system applies torque about the three principle axes to change its euler angles and angular velocities according to a PD control law as shown in (1).

$$\tau_{rw} = -K_p(e_{des}-e_{act})-K_d(\omega_{des}-\omega_{act}) \quad (1)$$

where $K_p$ and $K_d$ are the proportional and derivative controller gains, $e_{des}$ and $e_{act}$ are the desired and actual Euler angles, $\omega_{des}$ and $\omega_{act}$ are the desired and actual angular velocity of the spherical robot respectively. For ballistic hops on a horizontal surface, a proportional control command with $K_d$ equal to zero enables the robot to maintain constant Euler angle. Fig. 3 (top) shows the ballistic hopping capability of the spherical robot on a horizontal plane on Mars with acceleration due to gravity of 3.71 m/s$^2$.

However, for the robot to climb a vertical surface, a derivative control command with a desired angular velocity is used. Fig. 3(bottom) shows the ballistic hopping capability of the spherical robot on a vertical plane on Mars. The robot can climb a vertical distance of 1.27 m in 1.5 seconds while expending 5 grams of fuel and oxidizer. Fig. 4 shows the vertical distance that the robot is capable of climbing on the surfaces of Phobos, Ceres, Moon and Mars using 5 grams of fuel and oxidizer.

*C. Gripping Mechanism*

The performance of the multirobot system depends on the efficiency at which it can climb without slipping which in turn is highly dependent on its ability to grasp on a steeped natural surface. The gripping mechanism for each robot consists of an array of microspines embedded on a skin that wraps around its external surface.

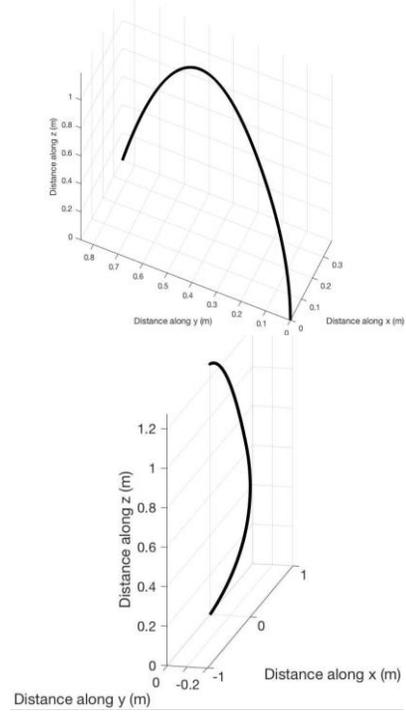

Fig. 3: Trajectory of Spherical robot performing a rocket-propelled ballistic hop on Horizontal surface (Top) and Vertical surface (Bottom).

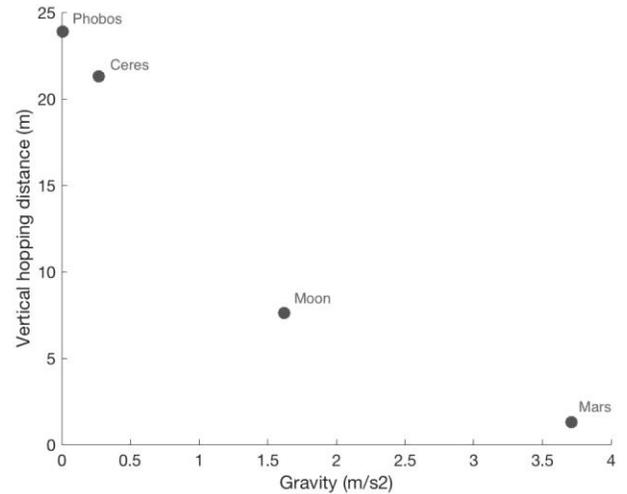

Figure 4: Vertical distance travelled in a single hop on Phobos, Ceres, Moon and Mars.

For the robot to grasp onto a wide variety of surfaces, the microspine array consists of a combination of microspines with a wide range of tip radius spread uniformly across the skin. Each microspine toe consists of a steel hook embedded in front of a rigid frame with elastic flexures acting as a suspension system as shown in Fig. 5 [14]. For a spine of tip radius $r_s$, it will engage to asperities of average radius $r_a$ such that $r_a \geq r_s$. The elastic flexures act as a suspension system and allow each hook to move relative to its neighbors. So, when an array of microspines are dragged along a rough surface, each toe is stretched and dragged to find a suitable asperity to grasp and share the overall load uniformly.

Engagement of the spine to asperities depend on the angle $\theta$ of the normal vector to the traced surface and is possible only if it is larger than some critical angle $\theta_{min}$ shown in Fig. 6. The angle $\theta_{min}$ depends on the angle at which the spines are loaded $\theta_{load}$, and coefficient of friction $\mu$, between the steel hook and the rocky surface as shown in (2). Hence, smaller spines with smaller tip radius $r_s$ are more effective at engaging to asperities on smooth surfaces. The maximum load that a spine can sustain is a function of the tensile stress of the hook and square of the radius of curvature of the spine tip and asperity as shown in (3) and (4) [12].

$$\theta_{min} = \theta_{load} + \cot^{-1} \mu \quad (2)$$

$$f_{max} = \left[\left(\pi \sigma_{max}/(1-2\mu)\right)^3 (1/2E^2)\right] R^2 \quad (3)$$

$$\frac{1}{R} = \frac{1}{r_s} + \frac{1}{r_a}\frac{1}{R} = \frac{1}{r_s} + \frac{1}{r_a} \quad (4)$$

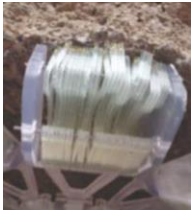

Fig. 5: Microspine toe securely gripping and hanging from a rocky surface [14].

Thus, as we decrease the tip radius of the hook, it can engage to smoother asperities but the load carrying capacity decreases. Hence, the design of the microspine skin has to be such that it can carry the load of the multirobot system and can engage onto a wide variety of rough surfaces. Each of the microspine toe consists an embedded elastic flexure mechanism that enables it to stretch parallel to the vertical surface under a load. Moreover each spine can stretch and drag relative to its neighbors to find a suitable asperity to grip. If a toe catches an asperity, neighboring toes will continue to slide down as the caught toe stretches and grip on a suitable asperity.

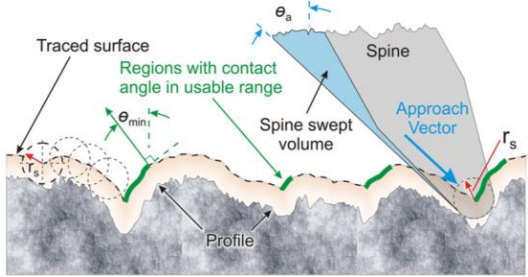

Fig. 6: Spine/surface interaction model [12].

### D. Sensing

For effective control and grasping, each robot must be capable of sensing its (inertial) angular orientation and angular velocity. Each robot employs an onboard inertial measurement unit (IMU) which consists of a 3-axis accelerometer and a gyroscope which are used to measure its orientation and angular velocities. Tactile sensors are employed to measure the contact forces and the onboard vision system with stereo cameras and laser range finders are used to measure the relative distance of each robot from a contact surface.

Aditionally for planning, each robot must be able to locate obstacles that are not feasible to climb. In addition, the robots need to find the best possible path and locate new griping points and possibly generate a description of their properties. The onboard stereo cameras and laser range finders are also used to locate obstacles and new gripping positions. Images are taken of the surroundings by the pinhole lens model and every point in a 3D space denoted by $M$ is transformed into a pixel $m$. The relationship between 3D point $M$ and its projected 2D point $m$ is shown in (5) [19]

$$sm' = A[RT]M' \quad (5)$$

where $s$ is a scaling factor, $R$ is a 3×3 rotation matrix, $T$ is a 3×1 translation vector and $A$ is a 3×3 matrix that describe the internal characteristics of the camera. By identifying and measuring nearest features points on an obstacle, we can calculate the obstacle-to-robot distance. Taking a panoramic view of the surroundings and locating the position of each obstacle and possible gripping points, each robot calculates its trajectory to determine where to hop next.

### IV. DYNAMIC SIMULATIONS AND EXECUTION

Climbing vertical or sloped surfaces in natural terrains such as cliffs is a challenging task as each climb is different. Climbing a vertical slope using a single robot is a risky task as it may slip and fall. However, a multirobot system can work cooperatively by being interlinked using spring tethers working like a team of mountaineers to systematically climb a slope. For a multirobot system, the planning problem consists of generating trajectories for each robot and moving the whole system up a vertical slope while maintaining equilibrium.

Our nominal system comprises of 4 robots interlinked together using 4 spring tethers in an "X" configuration as shown in Fig. 7. The connections between each robot and tether are made with ball and socket joints. The spring tether introduces three translational degrees of freedom in the system which allows each robot to translate with respect to the others. The ball-socket joint introduces three rotational degrees of freedom in the robot-tether and tether-tether connection, which allows each robot to hop with respect to others [5].

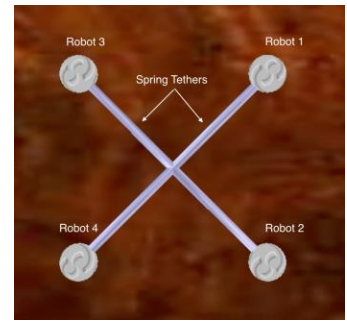

Fig. 7: Cliff climbing multirobot system

The terrain is modeled as a vertical rough surface using a form of a multivariate Weierstrass-Mandelbrot (W-M) function developed by Ausloos and Berman [20][21] as shown in (6) and (7).

$$z(x,y) = C \sum_{m=1}^{M} \sum_{n=0}^{n_{max}} \gamma^{(Ds-3)n} \left\{ \cos \phi_{m,n} - \cos \left[ \frac{2\pi \gamma^n (x^2+y^2)^{0.5}}{L} \cos \left( \tan^{-1} \left( \frac{y}{x} \right) - \frac{\pi m}{x} \right) + \phi_{m,n} \right] \right\} \quad (6)$$

$$C = L \left( \frac{G}{L} \right)^{Ds-2} \left( \frac{\ln \gamma}{M} \right)^{0.5} \quad (7)$$

Fig. 8 shows a section of the vertical terrain in μm scale. With the vertical surface defined, microspines of varying tip radius are uniformly distributed across the surface area of each robot. Each spine has a shaft diameter of 200-300 μm and a tip radius of 12-25 μm. The spines are loaded at angles $3.5^0 < \theta_{load} < 8^0$ degrees from the vertical surface. The coefficient of friction between stainless steel spine tips and rock varies from 0.15 to 0.25. Hence, the value of $\theta_{min}$ varies from $86.5^0$ to $81^0$ for an average $\theta_{load}$ of $5^0$. The approach angle $\theta_a$ is varied from $45^0$ to $65^0$ for the simulation. The maximum load that each spine can sustain per asperity is 1-2 N based on the radius of the tip and average radius of the asperities [12]. Considering 1.5 N as the average load each spine/asperity contact is capable of sustaining, the total gripping force is calculated for $N$ number of contact points as shown in (8).

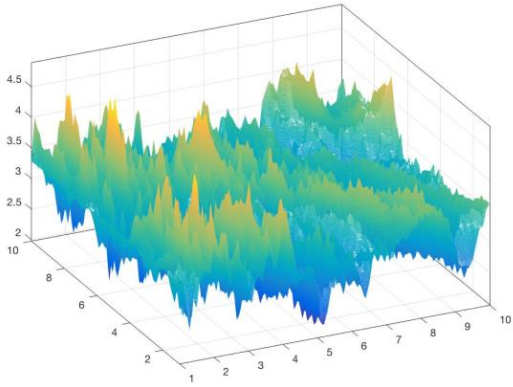

Fig. 8: Terrain of the vertical wall in μm scale

The forces acting on each robot while in contact with the vertical surface are the gripping contact force $F_c$, gravity $F_g$ and spring force $F_s$. The equilibrium condition for each robot is shown in (9). A robot looses surface contact and falls if the forces due to gravity $F_g$ and spring $F_s$ exceeds the contact force $F_c$.

$$F_c(\max) = 1.5N \quad (8)$$

$$F_c + F_g + F_s = 0 \quad (9)$$

Fig. 9 shows a Matlab 3D VRML dynamics simulation of a team of 4 robots climbing a vertical terrain. In Fig. 9(a) all robots are gripping onto the surface. Robot 1 disengages its grip and hops a distance $d$ forward and then grips again on the new location. When robot 1 hops, the other 3 robots are still engaged to the surface, hence if robot 1 fails to grip, slips or falls, the other 3 robots will be holding it up as they are anchored. Robot 1 continues to hop until it is able to grip onto the surface at a distance $d$ from its initial position. Similarly, Fig. 9(c)-9(e) shows robot 2,3 and 4 hopping and gripping on the surface until each robot engages into a new location. Fig. 9(e) shows the final configuration of the system after it climbed a distance $d$ up the slope. Fig. 9(f) shows the dynamics of the system when robot 4 hops but fails to grip onto the surface. Robot 4 slips, however, the other 3 robots anchored to the surface holds it up avoiding fall of the whole system.

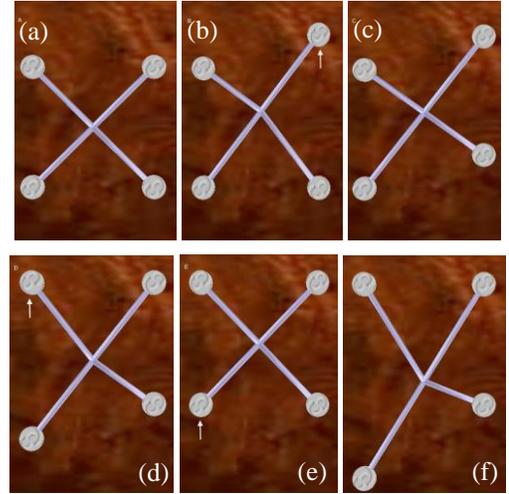

Fig. 9: Sequence of robot movement to climb a steep slope. Each robot hops up the slope, individually and in sequence and grips to the surface. Robot 4 fails to grip on the surface, however the multirobot system saves it from falling

Fig. 10 shows the change in position of each robot and the instantaneous center of the entire system with time. The initial position for robot 1,2,3 and 4 are (1.5,0,1.5), (1.5,0,0), (0,0,1.5) and (0,0,0) respectively. Each robot hops and grips one at a time resulting in change in position of the instantaneous center.

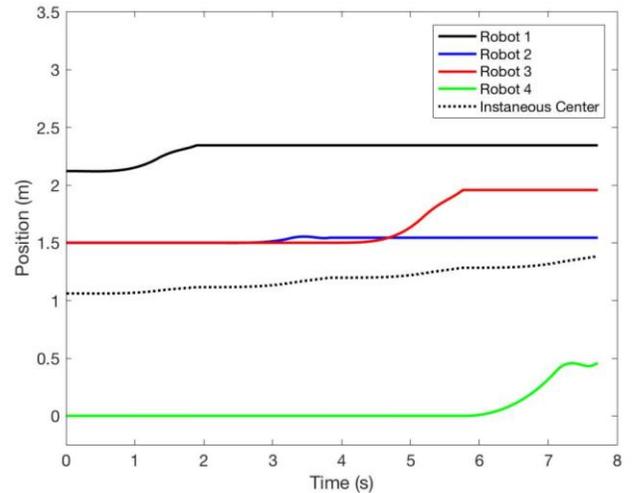

Fig. 10: Position of each robot during systematic climb.

Fig. 11 shows the z-coordinate of each robot and the instantaneous center when robot 1 fails to grip after hopping. All the robots successfully grip on the surface after the first hop. Robot 1 fails to grip after its second hop and slips down. The other 3 robots hold it up and allows it to hop again to attain the desired height.

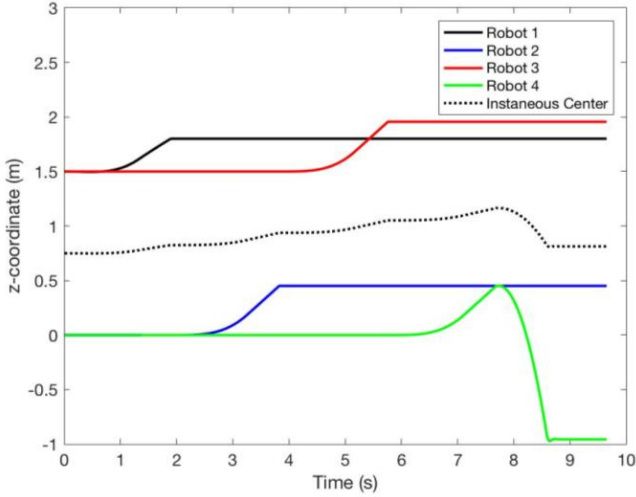

Fig. 11: The z-coordinate of each robot when robot 4 fails to grip and falls.

Further simulations were done for a 2, 3, 4, 5 and 6-robot systems to verify its feasibility. The simulations were performed for different number of spines able to grip per robot. Fig. 12 shows the probability of failure against the number of spines engaged to the surface per robot when one robot fails to grip for a 6, 5, 4, 3 and 2-robot system. Similarly, Fig. 13 shows the probability of failure against the number of spines engaged to the surface per robot (when two robots fail to grip). For each robotic systems, there is a critical number of spines per robot that needs to grip below which the probability of failure is 1 and above which it decreases significantly.

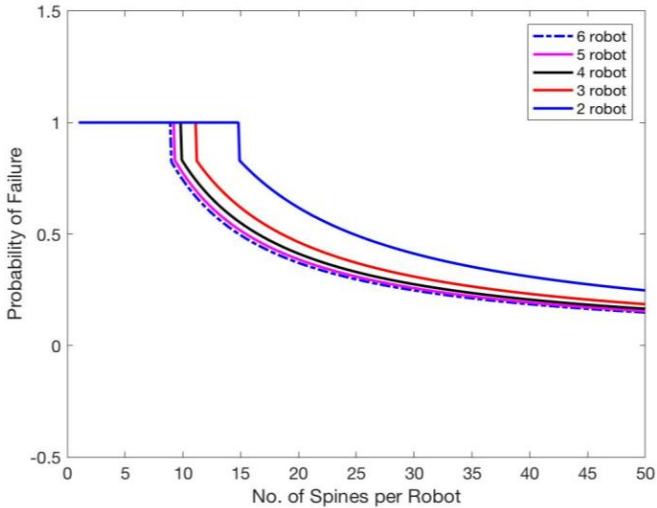

Fig. 12: Probability of failure when one robot fails to grip.

It is clear from the plots that when one robot fails, the minimum number of spines per robot needed to grip is 8 for a 6-robot system and 14 for a 2-robot system. Moreover, when two robots fails, the minimum number of spines per robot needed to grip increses to 11 for a 6-robot system and 22 for a 3-robot system.

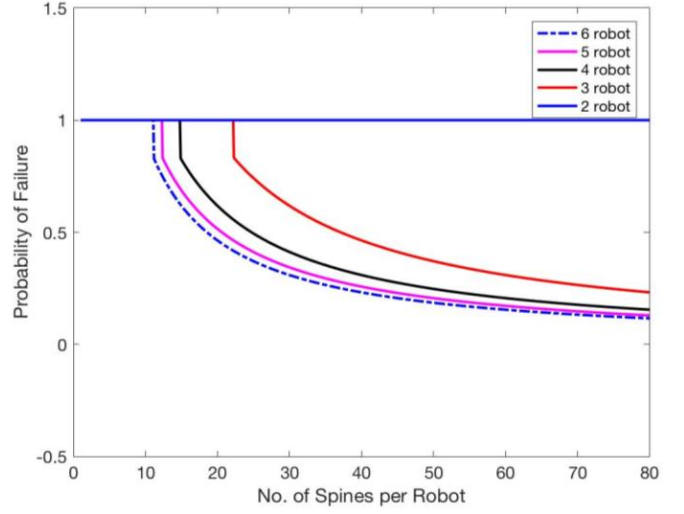

Fig. 13: Probability of failure when two robots fails to grip against number of spines per robot

Further studies were performed to verify the feasibility of the multirobot system with varying number of robots based on the number of robots that hops at once, required number of spines per robot to grip, maximum total distance travelled with 1 kg of propellant per robot, time required to travel the total distance, science data collected and communication links required among the robots. The number of spines per robot required to grip is propotional to the mass of the whole robotic system divided by the product of the minimum number of robots griping to the surface and the maximum load each spine can sustain as shown in (10). The maximun total distance travelled with 1 kg of propellant per robot is shown in (11) and the minimum time required to travel the total distance is shown in (12). The total science data collected is proportional to the area coverage of each robotic system as shown in (13) and the total communication links required for each system is proportional to the combination of the total number of robots as shown in (14). With increasing number of robots in each system, the total distance travelled by each system remains constant, the number of spines required to grip on the surface decreases while time required to climb a desired height, amount of science data generated and required communication links increases.

$$S \propto \frac{Nmg}{1.5(N-n)} \quad (10)$$

$$D \propto \frac{1000d}{5} \quad (11)$$

$$T \propto \frac{1000t}{5}\left(ceil\left(\frac{N}{n}\right)\right) \quad (12)$$

$$Area\ Coverage \propto N\pi r^2 - M\left(2r^2\cos^{-1}\left(\frac{D}{2r}\right) - \frac{D}{2}\sqrt{4r^2 - D^2}\right) \quad (13)$$

$$Comms\ Link \propto \frac{N!}{2!(N-2)!} \qquad (14)$$

In these conditions, $N$ is the total number of robots in each system, $n$ is the number of robots that hop at once, $m$ is the mass of each robot, $g$ is the acceleration due to gravity, $d$ is the distance that each robot hops at the expense of 5 grams of propellant, $t$ is the time required to hop a distance $d$, $r$ is the radial range of the onboard instrument in each robot, $D$ is the minimum seperation between each robot, $M$ is the nunber of overlaping regions in each robotic system and ceil($X$) is a function that rounds each element of $X$ to the nearest integer greater than or equal to that value. The raw values of each characteristic for each robotic systems are normalized to a range of 0 to 1, where 1 represents the fittest system and 0 represents an unfit system. The normalized fitness values for each characteristic is then multiplied to calculate the overall fitness metric of each robotic system. Fig. 14 shows the fitness metric of the overall system against the total number of robots in each system when one robot hops at a time. It is clear that the 4-robot system has the highest fitness while the 8-robot system has the lowest fitness.

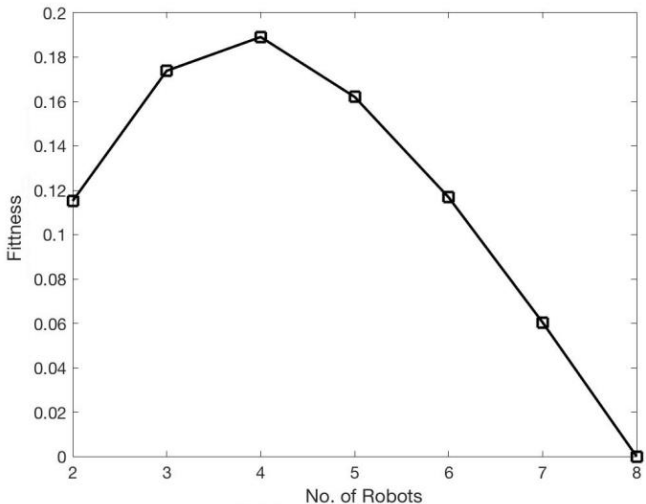

Fig. 14: Fitness of multi-robot system when one robot hops at a time.

Moreover, Fig. 15 shows the fitness values of the overall system versus the total number of robots in each system when two robots hop at a time. It is clear that the 6-robot system has the highest fitness while the 2-robot system has the lowest fitness (in a multirobot context).

## V. DISCUSSION

We propose the SphereX robotic platform for extreme environment exploration. It offers a compelling, practical solution that utilizes Commercial Off-The-Shelf (COTS) technologies to provide access to extreme environments not possible with current planetary rovers. Despite significant research in the field of COTS components, many conventional options are not practical for an off-world environment. Use of a miniature rocket system to propel the SphereX robot is simple, enabling hopping, short-flights and rolling. However, there are developmental challenges in miniaturizing the rocket thrusters.

Multiple SphereX robots are proposed for cooperative cliff climbing, where each robot is interlinked with tethers. In theory, this multirobot system is suitable for exploring cliff faces on Mars, the Moon, surfaces of asteroids and other planetary surfaces. With the propulsion unit, ADCS system and sensors such as IMUs incorporated in the system, each robot can perform ballistic hopping, short-flights and rolling. Moreover, with an array of microspines attached to each robot, the multirobot system can grip to any rough surface and then climb or crawl without the risk of falling from a cliff or flying off an asteroid. Climbing enables persistent access of the sloped surface.

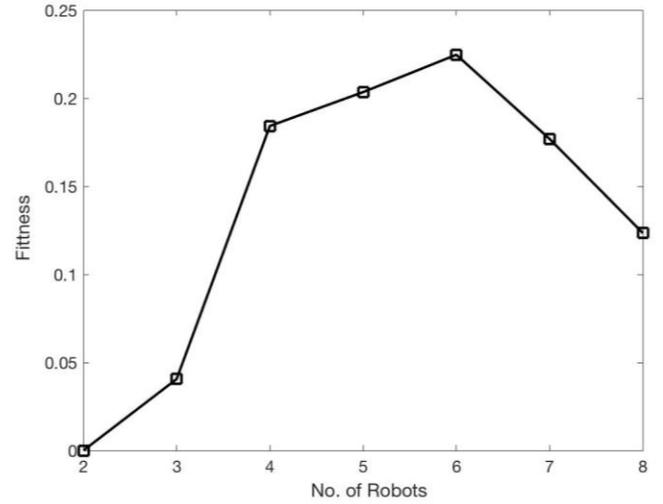

Fig. 15: Fitness of multi-robot system when two robots hop at a time.

We have performed detailed dynamics and control simulations for a 4-robot system. Each robot can hop a distance of 1.27 m in 1.5 seconds using 5 grams of RP1-$H_2O_2$ propellant in Mars. For the 4-robot system, when one robot hops at a time, the system performs four succesive hops to climb a distance of 1.27 m in apprroximately 6 seconds. Hence, with 1 kg of propellant available for each robot, the 4-robot system can climb a maximum distance of 254 m in 1,200 seconds. However, in case of failure of any robot to grip on the surface, the consumption of fuel and the time required to climb a certain height increases and the total distance the system is capable of climbing decreases significantly.

Similar analysis were done for a 2,3,5,6,7 and 8-robot system and the feasibility of each system were verified. As the number of robots increases in the system, the probability of failure for the system decreases as lesser number of spines per robot are required to grip on the surface but the time required to climb a certain height increases. Communication among the robots becomes complex as each robot needs to know the states of every other robot and the number of communication links required increases. However, with more number of robots climbing and reaching the desired science target, more science data can be collected. A comparative study among the robotic systems were performed and it is clear from Fig. 12 that the 4-robot system is superior than the other systems when one robot hops at a time. However, when two robots hop at a time, the 6-robot system is superior to the others. Increasing the number of robots improves the overall system performance, provided more of them can hop/climb simultaneously, while having enough robots anchored safely. These results hint at an optimal design principle for multi-robot climbing.

## VI. CONCLUSION

This paper presents a new spherical robot called SphereX that is small and utilizes unconventional mobility through hopping, flying or rolling to overcome low-gravity conditions. The SphereX robot is then used to introduce a multirobot system for cooperative cliff climbing and steeped surface exploration. The multirobot system is suitable for climbing cliff faces on the Moon, Mars and explore surfaces of low-gravity bodies like asteroids. The proposed system can withstand individual missteps, slips or falls by a robot during the climbing process. With the use of a propulsion system, reaction wheels and sensors like IMUs, the system can access hard to reach sites. The robots can simply fly to a location and land on to the side of a cliff before performing climbing up or down a few meters to reach a desired science target. We presented an overview of the hardware design of each SphereX robot. We also present the dynamics and control simulations for a single hopping robot with propulsion and ADCS system. Finally, the cliff climbing mechanism was simulated using multiple robots and multiple spring tethers. Our results suggests that increased parallelism of multiple robots interlinked in a chain attempting to hop to higher locations at once, while having others safely anchored optimizes climbing performance of the system. With these simulations, our efforts are now focused on realization and verification of these results on real robots.

ACKNOWLEDGMENTS

The authors would like to gratefully acknowledge Ravi Nallapu, Andrew Warren, Prof. Erik Asphaug, Prof. Mark Robinson and Dr. Stephen Schwartz for helpful discussions in the development of the SphereX robots.